\documentclass[final]{cvpr}

\usepackage{amsmath,amsfonts,bm}

\def\eqref#1{equation~\ref{#1}}

\def\1{\bm{1}}

\DeclareMathAlphabet{\mathsfit}{\encodingdefault}{\sfdefault}{m}{sl}
\SetMathAlphabet{\mathsfit}{bold}{\encodingdefault}{\sfdefault}{bx}{n}

\DeclareMathOperator*{\argmax}{arg\,max}

\usepackage{times}
\usepackage{epsfig}
\usepackage{graphicx}
\usepackage{amsmath}
\usepackage{amssymb}
\usepackage{algorithm} 
\usepackage{algorithmic} 
\usepackage{multirow} 
\usepackage{amssymb} 
\usepackage{xcolor}
\usepackage{booktabs} 
\usepackage{subfigure}	
\usepackage{bbding}
\usepackage{pifont}
\usepackage{wasysym}
\usepackage{bm}
\usepackage[accsupp]{axessibility} 

\newtheorem{Property}{Property} 

\newtheorem{Theorem}{Definition} 

\newtheorem{Remark}{Remark}

\usepackage[pagebackref=true,breaklinks=true,colorlinks,bookmarks=false]{hyperref}

\def\cvprPaperID{01744} 
\def\confYear{CVPR 2022}

\begin{document}

\title{Weakly Supervised Object Localization as Domain Adaption}

\author{Lei Zhu$^{1,3,5}$ 
~ Qi She$^{2}$ 
~ Qian Chen$^{1,3,5}$ 
~ Yunfei You$^{1,3,5}$ 
~ Boyu Wang$^{4}$ 
~ Yanye Lu$^{*,1,5}$ \\
{\tt\small zhulei@stu.pku.edu.cn, \tt\small yanye.lu@pku.edu.cn}
\and $^{1}$Institute of Medical Technology, Peking University 
\and $^{2}$Bytedance AI Lab 
\and $^{3}$Department of Biomedical Engineering, Peking University 
\and $^{4}$University of Western Ontario
\and $^{5}$Institute of Biomedical Engineering, Peking University Shenzhen Graduate School 
}

\maketitle

\begin{abstract}
Weakly supervised object localization (WSOL) focuses on localizing objects only with the supervision of image-level classification masks. Most previous WSOL methods follow the classification activation map (CAM) that localizes objects based on the classification structure with the multi-instance learning (MIL) mechanism. However, the MIL mechanism makes CAM only activate discriminative object parts rather than the whole object, weakening its performance for localizing objects. To avoid this problem, this work provides a novel perspective that models WSOL as a domain adaption (DA) task, where the score estimator trained on the source/image domain is tested on the target/pixel domain to locate objects. Under this perspective, a DA-WSOL pipeline is designed to better engage DA approaches into WSOL to enhance localization performance. It utilizes a proposed target sampling strategy to select different types of target samples. Based on these types of target samples, domain adaption localization (DAL) loss is elaborated. It aligns the feature distribution between the two domains by DA and makes the estimator perceive target domain cues by Universum regularization. Experiments show that our pipeline outperforms SOTA methods on multi benchmarks. Code are released at \url{https://github.com/zh460045050/DA-WSOL_CVPR2022}.
\end{abstract}

\section{Introduction}
Weakly supervised object localization (WSOL), learning the location of objects in images using only the image-level classification mask to supervise, relaxes the requirement for the dense annotation such as pixel-level segmentation masks or bounding boxes for the training process, which saves much manual labor for annotation and attract extensive attention in recent years~\cite{CAM, ADL, PSOL, CAAM, I2C, PAS, IVR}. 
\begin{figure}[!htp]
\centering
\includegraphics[width=0.49\textwidth]{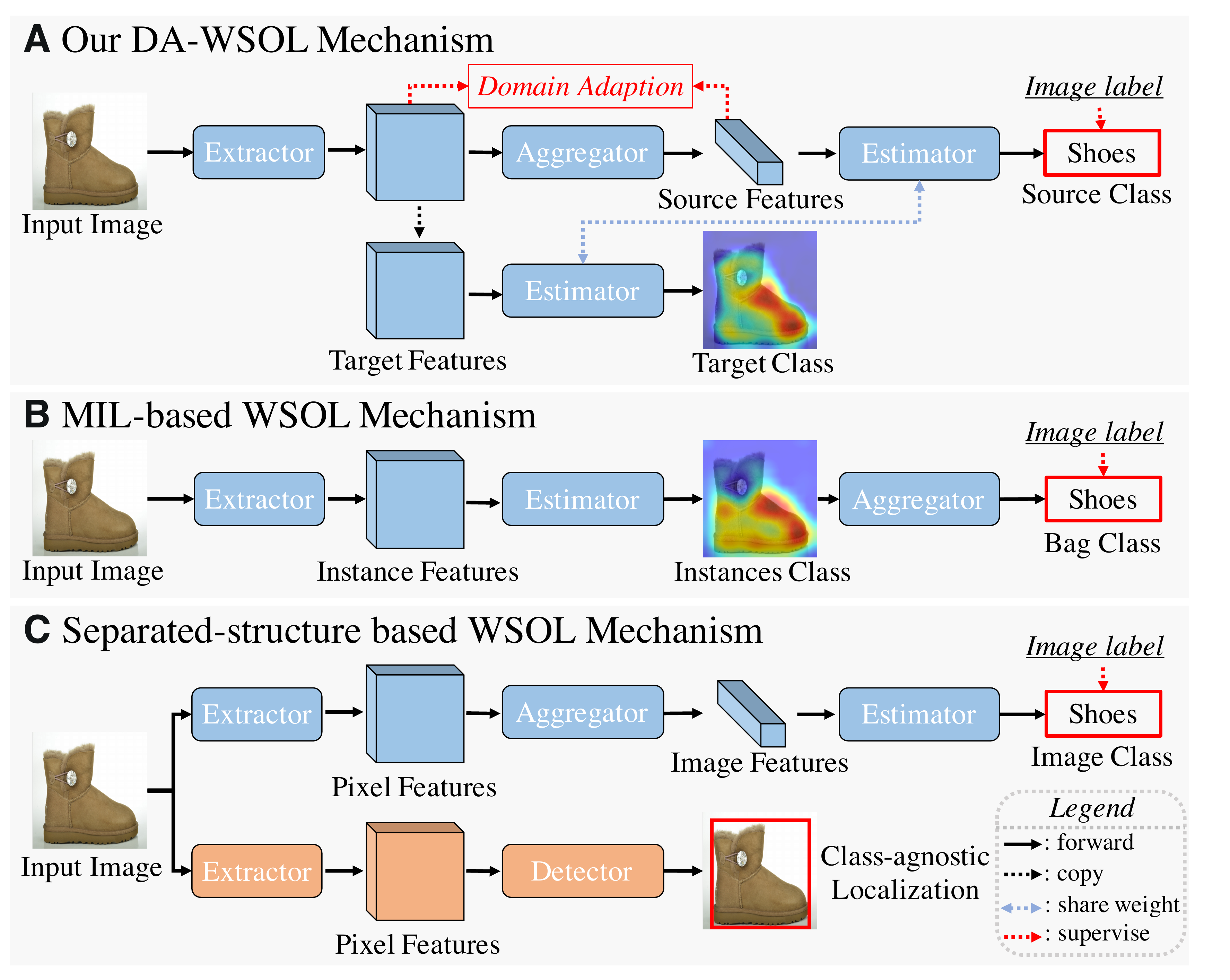}
\caption{The comparison of DA-WSOL and other mechanisms.}
\label{fig:intro}
\end{figure}

The most well-known WSOL method is the classification activation map (CAM)~\cite{CAM}, which utilizes the classification structure to generate the localization score. As indicated in Fig.~\ref{fig:intro} \textbf{\sf{B}}, it theoretically serves as classification under multi-instance learning (MIL)~\cite{MIL}, where each image represents a bag whose label is determined by instances it contains~\cite{ACOL}, \textit{i.e.} the pixels/patches. However, MIL focuses more on the accuracy of bags (images) rather than the instances (pixels), which makes CAM only discern the most discriminative parts but not the whole object. For activating more objects parts, different technics are adopted to enhance CAM, \textit{e.g.} data augmentations~\cite{CUTMIX, HAS, ACOL, ADL}, novel network structures~\cite{ADL, SPG, CAAM, I2C}, or post-processes~\cite{PAS, IVR}. Though these methods somewhat alleviate this problem, they still follow MIL that basically causes the incomplete activation of objects and weakens the performance.


From another perspective, CAM is also the same as training an estimator to classify the image-level feature under the supervision of image-level masks. This well-trained estimator is then projected onto the pixel-level feature to generate the pixel-level localization scores in the test processing~\cite{CAM}. By viewing image-level and pixel-level features as the features respectively extracted from source and target domain, the goal of WSOL is consistent with the domain adaption (DA) task, \textit{i.e.} forcing the estimator trained on the source (image) domain to perform well on the target (pixel) domain. Thus, if DA methods can assist WSOL in aligning the distribution of these two domains, the estimator can avoid overfitting the source domain, \textit{i.e.} only activating the discriminative parts of objects.

Inspired by this, our paper elaborates a DA-WSOL pipeline that helps to better engage the DA approaches into WSOL to enhance the performance. Fig.~\ref{fig:intro} visually generalizes our novel DA-WSOL pipeline and its difference from other mechanisms of WSOL. Compared with the MIL mechanism, our method adopts DA to align the feature distribution between source and target domain, enhancing the accuracy of the estimator when projected onto the pixel features. Moreover, both classification and localization scores can be obtained with a single end-to-end trained structure. This trait makes our method more concise than another type of WSOL mechanism~\cite{PSOL, GC} shown in Fig.~\ref{fig:intro} \textbf{\sf{C}}, which requires training multi additional stages to generate class-agnostic region-of-interests (ROI) when localizing objects.

In addition, the specificities of domain adaption in WSOL are also considered by the proposed DA-WSOL pipeline. Specifically, the target domain of WSOL is more complex constructed, which has a much larger scale than the source domain and contains samples with the source-unseen label, \textit{i.e.} the background locations. Thus, a target sampling strategy is also proposed for our DA-WSOL pipeline to effectively select the source-related target samples and the unseen target samples. The two types of samples are then fed into a proposed domain adaption localization (DAL) loss, where the former type is utilized to solve the sample imbalance between two domains, and the latter is viewed as the Universum~\cite{UNIVERSUM} to perceive target cues.

In a nutshell, our contributions are fourfold:
\begin{itemize}
\item {Our work is the first to model WSOL as a DA task and designs a pipeline to assist WSOL by DA methods.}
\item {A DAL loss is proposed to align the feature distribution of source and target domain in the WSOL scenario.}
\item {A target sampling strategy is proposed to select different types of representative target samples.}
\item{Extensive experiments show our DA-WSOL outperforms SOTA methods on multiply WSOL benchmarks.}
\end{itemize}

\section{Related Works}
\subsection{Multi-instance learning based WSOL}
MIL-based WSOL methods train a classification network in a MIL manner to ensure that this network can be utilized to localize pixels with certain classes. Zhou~\textit{et al.}~\cite{CAM} proposed the CAM, which utilizes the classification structure to localize objects by projecting the classifier onto the pixel-level feature. Zhang~\textit{et al.}~\cite{ACOL} proved its mechanism is equal to the MIL and improved it by utilizing two adversarial classifiers to catch complementary object parts. For a similar purpose, Singh~\textit{et al.}~\cite{HAS} randomly hid the patches of images to force the classification structure to focus more parts of objects. Junsk~\textit{et al.}~\cite{ADL} then enhanced it by directly erasing discriminative spatial positions on the feature map. Yun~\textit{et al.}~\cite{CUTMIX} replace patches of an image by its of other images to augment the training samples. Except for the data augmentations, novel network structures were also explored to enhance the CAM. Typically, Zhang~\textit{et al.}~\cite{SPG} generated coarse pixel-wise masks based on different stages of the extractor and used them as additional supervision to force the later stages. Zhang~\textit{et al.}~\cite{I2C} also adopted the siamese network to ensure stochastic consistency for images with the same class in a batch. Unlike these methods that follow MIL mechanism, our method align pixel-level and image-level features with DA approches to assist WSOL task.

\subsection{Separated-structure based WSOL}
Instead of viewing WSOL as the image classification in the MIL manner, Zhang~\textit{et al.}~\cite{PSOL} suggested splitting WSOL into two independent tasks: the class-agnostic object localization and the object classification. They proposed a pseudo-supervised object localization (PSOL) method with three stages respectively for region proposal, bounding box regression, and object classification. Lu~\textit{et al.}~\cite{GC} attempted to generate ROI with different geometry and adopted a generator to produce class-agnostic binary masks rather than only bounding boxes. Zhang~\textit{et al.}~\cite{SEM} used the classification stage to obtain both classification results and localization seeds, which are then used by the localization stage to generate class-agnostic localization maps. More recently, Meng~\textit{et al.}~\cite{FAM} explored jointly optimizing localization and classification to pursue better results. Compared with them, our method can obtain both classification and localization results with only one stage.

\subsection{Domain Adaption}
Domain adaption aims at learning discriminative models in the presence of the shift between distributions of the training and test samples. Some DA methods focus on catching domain-invariant features with deep neural networks by minimizing the maximum mean discrepancy (MMD)~\cite{MMD, MMD2, MMD3}. Long~\textit{et al.}~\cite{JAN} enhanced the MMD by considering the joint distributions for the feature of different neural network stages. Kang~\textit{et al.}~\cite{CAN} and Zhu~\textit{et al.}~\cite{DSAN} further took the class label of samples into account to measure both intra-class and inter-class discrepancies. Instead of measuring the distribution discrepancy, some methods also adopt adversarial learning to confuse a well-trained domain classifier when learning domain-invariable features. Typically, Ganin~\textit{et al.}~\cite{DANN} added a gradient reversal layer right before the domain classifier to enable adversarially training the domain adaption network in an end-to-end manner. Pei~\textit{et al.}~\cite{MADA} further considered fine-grained alignment of different data distributions by adopting multiply domain classifiers with gradient reversal layer. Yu~\textit{et al.}~\cite{DAAN} subdivided those classifiers into global and sub-domain classifiers to learn the relationship between the marginal and conditional distributions. Different from those methods, our work focus on adopting DA to better assist WSOL task.

\section{Method}
This section firstly revisits WSOL and provides a novel perspective that WSOL can be modeled as a DA task with specific properties. Then, the proposed DAL loss is introduced, which considers the effects of different types of target samples based on the specificities of WSOL. Next, we detail the target sampling strategy that assigns the target samples into the three types. Finally, the entire structure and workflow of our DA-WSOL pipeline are introduced in detail. Note that the meanings of all the key symbols are also given in our Appdenix for clarity.

\subsection{Revisiting the WOSL}

Given an input image $\bm{X} \in \mathbb{R}^{3 \times N}$, the object localization task aims at discerning whether a position $\bm{X}_{:, i}$ belongs to the object of a certain class $k$, where $N$ is the number of pixels in the image. For this purpose, a feature extractor $f(\cdot)$ and a score estimator $e(\cdot)$ are learned to extract the pixel-level features $\bm{Z}=f(\bm{X}) \in \mathbb{R}^{C \times N}$ and estimate the localization score $\bm{Y}^{*} = e(\bm{Z}) \in \mathbb{R}^{K \times N}$, respectively. For fully supervised object localization, the pixel-level localization mask $\bm{Y} \in \mathbb{R}^{K \times N}$ is adopted as supervision for $\bm{Y}^{*}$ to learn $f(\cdot)$ and $e(\cdot)$. Note that, the element $\bm{Y}_{k, i}$ identifies whether or not pixel $i$ belongs to the object of the class $k$.

As for WSOL, only the image-level classification mask $\bm{y} = \big(\max(\bm{Y}_{0, :}), \max(\bm{Y}_{1, :}), ..., \max(\bm{Y}_{K-1, :})\big) \in \mathbb{R}^{K \times 1}$ is available for the whole training process. Thus, an additional aggregator $g(\cdot)$ is added between the feature extractor and the score estimator to aggregate the pixel-level feature into image-level, \textit{i.e.} $\bm{z} = g(\bm{Z}) \in \mathbb{R}^{C \times 1}$. This image-level feature is then fed into the score estimator to generate the image-level classification score $\bm{y}^{*} =e(\bm{z})\in \mathbb{R}^{K \times 1}$. Profited by this image-level score, the image-level mask $\bm{y}$ can supervise the training process with the classification loss, such as the cross-entropy between $\bm{y}^{*}$ and $\bm{y}$. While in the test time, this estimator is projected back onto the pixel-level feature $\bm{Z}$ to predict the localization scores $\bm{Y}^{*}$.

\begin{figure*}
\centering
\includegraphics[width=0.96\textwidth]{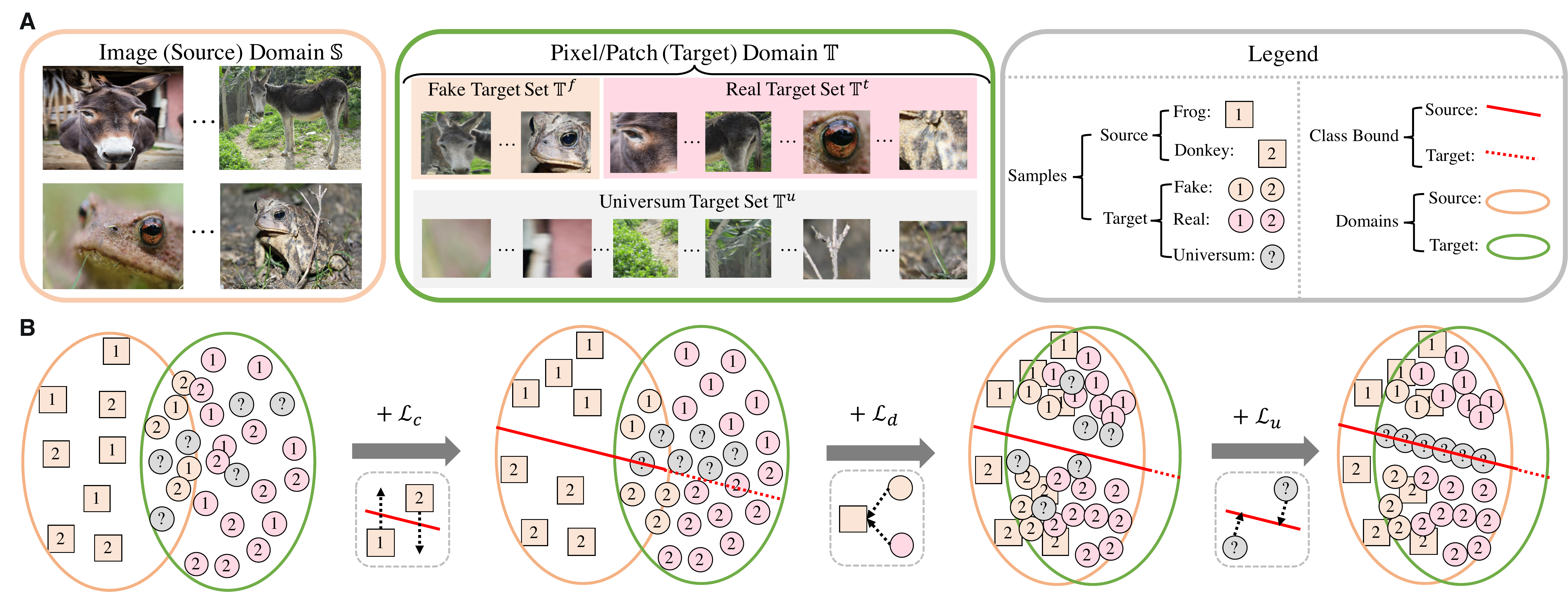}
\caption{The mechanism of our proposed DA-WSOL (best view in color). \textbf{\sf{A}}. The visualization of the source domain, target domain, and different types target samples. \textbf{\sf{B}}. The effects shown when sequentially adding our designed loss function $\mathcal{L}_{c}$, $\mathcal{L}_{d}$, and $\mathcal{L}_{u}$. }
\label{fig:subset}
\end{figure*}

\subsection{Modeling WSOL as Domain Adaption}

By mapping all input images into two feature spaces represented by $\bm{s} = \bm{z} = (g \cdot f)(\bm{X})$ and $\{ \bm{t}_{1}, \bm{t}_{2}, ..., \bm{t}_{N}\} = \{ \bm{Z}_{1, :}, \bm{Z}_{2, :}, ..., \bm{Z}_{N, :} \} = f(\bm{X})$, two feature sets $\mathbb{S}:\{\bm{s}\}$ and $\mathbb{T}:\{\bm{t}\}$ can be constructed, where we call $\mathbb{S}$ and $\mathbb{T}$ as the source (image) domain and target (pixel) domain for clarity. The corresponding label sets of $\mathbb{S}$ and $\mathbb{T}$ are defined as $\mathbb{Y}^{s}:\{\bm{y}^{s}=\bm{y}\}$ and $\mathbb{Y}^{t}:\{\bm{y}^{t}=\bm{Y}_{:, i}\}$, respectively. Under this view, WSOL can be seen as fully-supervised training the score estimator $e(\cdot)$ on the image domain $\mathbb{S}$ with label set $\mathbb{Y}^{s}$, and then taking inference on the pixel domain $\mathbb{T}$ to estimate the localization mask $\mathbb{Y}^{t}$ in the test process. This process is a typical setting of the DA task, which aims at learning a discriminative model in the presence of the feature shift between training and test process~\cite{DAAN}. Thus, WSOL is equal to solving the following DA task:
\begin{Theorem}
Given an image set $\mathbb{X}=\{\bm{X}^{1}, ..., \bm{X}^M \}$, the source domain and target domain are defined as:
\begin{equation} 
\left\{
\begin{aligned}
&\mathbb{S} : \{\bm{s} = (g \cdot f)(\bm{X}^{m}) ~|~ m \in [1, M]\} \\
& \mathbb{T} : \{\bm{t} = f(\bm{X}^{m})_{:, i} ~|~ i \in [1, N], m \in [1, M]\}
\end{aligned}
\right. .
\end{equation}

\noindent WSOL aims at minimizing the target risk without accessing the target label set $\mathbb{Y}^{t}$. It serves as a multi-task problem: 1) minimizing source risk between $e(\mathbb{S})$ and $\mathbb{Y}^{s}$. 2) minimizing domain discrepancy between $\mathbb{S}$ and $\mathbb{T}$.
\label{prop:def}
\end{Theorem}

Based on the perspective of Theorem.~\ref{prop:def}, the object of our proposed DA-WSOL pipeline can be formulated as:
\begin{equation}
\mathcal{L}(\mathbb{S}, \mathbb{Y}^{s}, \mathbb{T} ) = \mathcal{L}_{c}(e(\mathbb{S}), \mathbb{Y}^{s}) + \mathcal{L}_{a}(\mathbb{S}, \mathbb{T})~~,
\end{equation}

\noindent where $\mathcal{L}_{c}(\cdot)$ is the classification loss that supervises the accuracy on the source domain. $\mathcal{L}_{a}(\cdot)$ is the adaption loss minimizing the discrepancy between the $\mathbb{S}$ and $\mathbb{T}$. This additional term forces $f(\cdot)$ and $g(\cdot)$ to learn domain-invariant features between the source (image) domain and target (pixel) domain, which helps the source trained estimator $e(\cdot)$ also perform well on target samples. Thus, more object locations can be activated in the test processing.
 
\begin{Remark}
There are also some specific properties for WSOL that do not exist in the traditional DA task. These properties weaken the applicability for directly implementing $\mathcal{L}_{a}(\cdot)$ as the existing DA methods~\cite{MMD, JAN, CAN, DSAN, DANN, DAAN}.
\end{Remark}

\begin{Property}
Target domain $\mathbb{T}$ contains samples that do not belong to any object classes of source domain $\mathbb{S}$, \textit{i.e.} the background locations. Aligning features of these samples with the source domain will hurt the performance.
\label{prop:background}
\end{Property}

\begin{Property}
The number of samples in the source domain is much less than the target domain in WSOL, \textit{i.e.} $|\mathbb{S}| = |\mathbb{T}| / N$. The insufficient samples cause difficulties when perceiving the source distribution during training.
\label{prop:unbalance}
\end{Property}

\begin{Property}
The discrepancy between the source domain and target domain is attributed to the aggregation function $g(\cdot)$ rather than entirely unknown as traditional DA tasks. This property can be used as a prior when aligning the feature distribution of source and target domain.
\label{prop:prior}
\end{Property}

\subsection{Domain Adaption Localization Loss}
To consider these properties, a domain adaption localization (DAL) loss is elaborated as $\mathcal{L}_{a}(\cdot)$. As shown in Fig.~\ref{fig:subset} \textbf{\sf{A}}, our DAL loss further divides the target set $\mathbb{T}$ into three subsets based on the above properties: 1) the fake target set $\mathbb{T}^{f}:\{\bm{t}^{f}\}$ contains target samples that are highly correlated to the source domain. 2) the Universum set $\mathbb{T}^{u}:\{\bm{t}^{u}\}$ contains target samples whose label is unseen in source domain, \textit{i.e.} background locations.;  3) the real target set $\mathbb{T}^{t}:\{\bm{t}^{t}\}$ contains target samples that do not belong to $\mathbb{T}^{f}$ and $\mathbb{T}^{u}$.

Specifically, based on Property.~\ref{prop:prior}, the source domain is constructed by operating spatial aggregation on target samples. The distribution of some target samples that have high importance on the aggregation functions $g(\cdot)$ is similar to the source domain, for example the patches of frog or donkey head in Fig.~\ref{fig:subset} \textbf{\sf{A}}. These samples are contained in $\mathbb{T}^{f}$ and then used as a supplement to estimate the distribution of the source domain $\mathbb{S}$ that has insufficient samples as discussed in Property.~\ref{prop:unbalance}. Moreover, to solve the unmatched label space between the source and target domain as discussed in Property.~\ref{prop:background}, $\mathbb{T}^{u}$ is utilized to contain samples that have the source-unseen label, for example the patches of ground or grass in Fig.~\ref{fig:subset} \textbf{\sf{A}}. It ensures that $\mathbb{T} - \mathbb{T}^{u}$ has the same label space as the source domain. Finally, by purifying $\mathbb{T}^{f}$ and $\mathbb{T}^{u}$ from $\mathbb{T}$, other samples construct $\mathbb{T}^{t}$ that is used to estimate the distribution of the target domain.

Based on these three subsets of the target domain, the DAL loss is defined to minimize the domain discrepancy:
\begin{equation}
\mathcal{L}_{a}(\mathbb{S}, \mathbb{T}) = \mathcal{L}_{DAL}(\mathbb{S}, \mathbb{T}) = \lambda_{1}\mathcal{L}_{d}(\mathbb{S}\cup\mathbb{T}^{f}, \mathbb{T}^{t}) + \lambda_{2}\mathcal{L}_{u}(\mathbb{T}^{u})~~, 
\label{eq:DAL}
\end{equation}

\begin{figure*}
\centering
\includegraphics[width=0.95\textwidth]{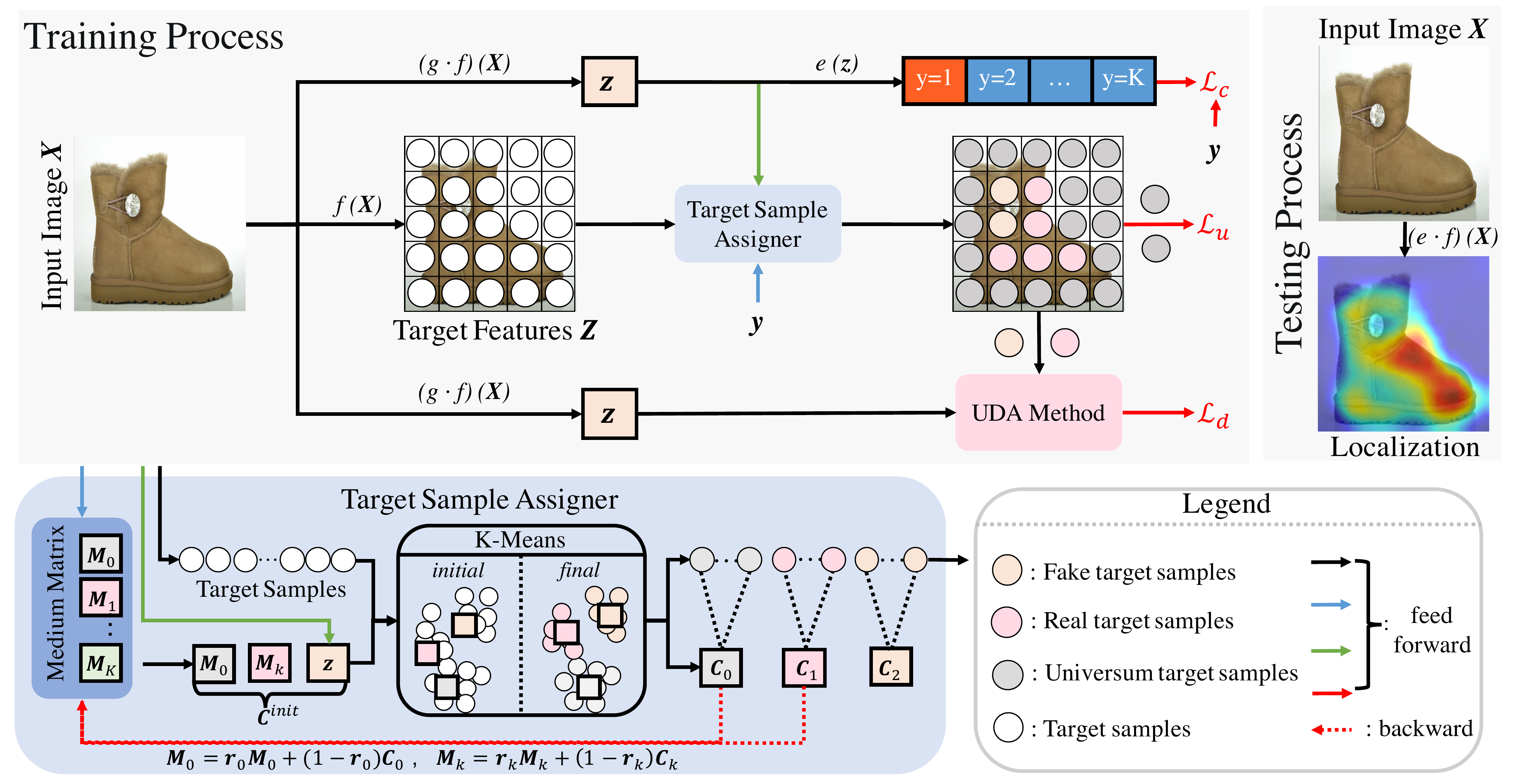}
\caption{The total structure and work flow of the proposed DA-WSOL pipeline (best view in color).}
\label{fig:pipeline}
\end{figure*}

\noindent where $\lambda_{1}, \lambda_{2}$ are two parameters, $\mathcal{L}_{d}(\cdot)$ is the domain adaption loss, and $\mathcal{L}_{u}(\cdot)$ is the Universum regularization~\cite{UNIVERSUM}. In detail, the domain adaption loss $\mathcal{L}_{d}(\cdot)$ can be implemented as unsupervised domain adaption (UDA) approaches~\cite{MMD, DANN} to align the feature distributions between the two domains without accessing target labels. As visualized in Fig.~\ref{fig:subset} \textbf{\sf{B}}, adding $\mathcal{L}_{d}(\cdot)$ can tighten the source domain (peach circle) and the target domain (green circle), making the source-trained estimator also perform better for the target samples. 

In addition, $\mathcal{L}_{u}(\cdot)$ adopts the mechanism of Universum~\cite{UNIVERSUM} that uses target samples with the source-unseen label ($\mathbb{T}^{u}$) to enhance the performance on the target set. It is implemented as feature-based $l_{1}$ regularization:
\begin{equation}
\mathcal{L}_{u}(\mathbb{T}^{u}) = \sum_{\bm{t}^{u}_{i} \in \mathbb{T}^{u}} |\bm{t}^{u}_{i}|~~,
\label{eq:Universum}
\end{equation}

\noindent which minimizes the feature strength of Universum samples. As visualized in Fig.~\ref{fig:subset} \textbf{\sf{B}}, adding $\mathcal{L}_{u}$ pushes the decision boundary into Universum samples, which makes the decision boundary also concerns target cues~\cite{UNIVERSUM}. Moreover, this regularization also reduces the domain discrepancy attributed to $g(\cdot)$, because it eliminates noises caused by Universum samples when generating source feature.

\subsection{Target Sampling Strategy}
The calculation of DAL loss requires samples of the three subsets on the target domain. Thus, a target sampling strategy is proposed for selecting the representative samples of the these subsets from target features of a certain input image. The core of this strategy is a target sample assigner (TSA) shown in Fig.~\ref{fig:pipeline}, which contains a cache matrix $\bm{M} \in \mathbb{R}^{\ C \times  (K+1) }$. In detail, $\bm{M}_{:, 0}$ represents the anchor of $\mathbb{T}^{u}$ and is initialized as the zero vector before training. Other column vectors, for example $\bm{M}_{:, k+1}$, represent the anchor of $\mathbb{T}^{t}$ for a certain class $k$. When the image of class $k$ is first accessed, $\bm{M}_{:, k+1}$ is initialized by adding a small random offset $\epsilon$ on its source feature $\bm{z}$, \textit{i.e.} $\bm{M}_{:, k+1} = \bm{z} + \epsilon$.

Profited by the cache matrix $\bm{M}$, in the forward pass of the training process, the TSA can provide the anchors of the $\mathbb{T}^{u}$ and $\mathbb{T}^{t}$ based on the image-level mask $\bm{y}$:
\begin{equation} 
\bm{a}^{u} = \bm{M}_{:, 0}~~,~~
\bm{a}^{t} = \bm{M}_{:, k+1}~~, ~~ k=\argmax(\bm{y})~~,
\end{equation}

\noindent where $k$ is the class index. $\bm{a}^{u}, \bm{a}^{t} \in \mathbb{R}^{ C \times 1}$ are the anchors of $\mathbb{T}^{u}$ and $\mathbb{T}^{t}$, respectively. These two anchors are then combined with the source feature $\bm{z}$ to form the initial center of the three subsets, \textit{i.e.} $\bm{C}^{init} = \{\bm{a}^{u}, \bm{a}^{t}, \bm{z} \} \in \mathbb{R}^{ C \times 3}$.

Next, based on $\bm{C}^{init}$, the three-way K-means clustering is operated on the target samples, \textit{i.e.} the column vectors of $\bm{Z}$, to assign them a cluster label $y^{c}_{i} \in \{0, 1, 2\}$. Finally, $n$ samples are randomly sampled for each subset based on the cluster labels to calculate the DAL loss:
\begin{equation} 
\left\{
\begin{aligned}
& \mathbb{T}^{u} : \{ \bm{Z}_{:, i} ~|~ y^{c}_{i} = 0 \} \\
& \mathbb{T}^{t} :  \{ \bm{Z}_{:, i} ~|~ y^{c}_{i} = 1 \} \\
& \mathbb{T}^{f} :  \{ \bm{Z}_{:, i} ~|~ y^{c}_{i} = 2 \}
\end{aligned}
\right. ~~.
\end{equation}

In the backward process, the final cluster center $\bm{C}$ outputted by K-means clustering is utilized to optimize $\bm{M}$ gradually during the training process:
\begin{equation} 
\bm{M}_{k, :} = \left\{
\begin{aligned}
& \bm{r}_{0} \bm{M}_{:, 0} + (1-\bm{r}_{0}) \bm{C}_{:, 0} ~, ~k = 0 \\
& \bm{r}_{k} \bm{M}_{:, k} + (1-\bm{r}_{k}) \bm{C}_{:, 1} ~, ~k \neq 0, \bm{r}_{k} \neq 0 \\
& \bm{C}_{:, 1} ~, ~k \neq 0, \bm{r}_{k} = 0, ||\bm{C}_{:, 2} - \bm{z}|| \leq ||\bm{C}_{:, 1} - \bm{z}|| \\
& \bm{C}_{:, 2} ~, ~k \neq 0, \bm{r}_{k} = 0, ||\bm{C}_{:, 2} - \bm{z}|| > || \bm{C}_{:, 1} - \bm{z}||
\end{aligned}
\right.
\label{eq:update}
\end{equation}

\noindent where $\bm{r} \in \mathbb{R}^{1 \times (K+1)}$ contains the update ratio.  $\bm{r}_{k}$ is implemented as reciprocal for the number of the passed images with class $k$. Based on Eq.~\ref{eq:update}, $\bm{M}$ can approximate the centroid of the $\mathbb{T}^{u}$ and $\mathbb{T}^{t}$, which enhances the accuracy of the anchor it provides. Note that, if the anchor of target domain initialized by $\bm{z} + \epsilon$, \textit{i.e.} $\bm{r}_{k} = 0$, we choose the cluster center that has large distance to $\bm{z}$ as the updated anchor.

\subsection{DA-WSOL Pipeline}
\label{sec:pipeline}
The proposed target sampling strategy can be easily engaged into current WSOL methods to train them with DAL loss. It acts as a connection for our DA-WSOL pipeline to improve the performance of the WSOL method with DA approaches. Fig.~\ref{fig:pipeline} shows the entire structure of the proposed DA-WSOL pipeline. Specifically, in the training process, the source domain $\mathbb{S}$ and the target domain $\mathbb{T}$ for a batch of images are firstly generated by feeding the input image into the feature extractor $f(\cdot)$ and the feature aggregator $g(\cdot)$ of a certain WSOL method. Here, we follow the CAM that implements  $f(\cdot)$ as the classification backbone (ResNet~\cite{RESNET}, InceptionV3~\cite{INCEPTION}) and adopts global average pooling (GAP)~\cite{GAP} as $g(\cdot)$. Moreover, the estimator $e(\cdot)$ implemented by the fully-connected layer is operated on the sample of the source domain $\mathbb{S}$ to generate the image-level classification score, which is supervised by their image-level mask $\bm{y}$ with cross-entropy:
\begin{equation}
\mathcal{L}_{c}(e(\mathbb{S}), \mathbb{Y}^{s})  = \sum_{ (\bm{s}_{i}, \bm{y}_{i}) \in (\mathbb{S}, \mathbb{Y}^{s})}\mathcal{L}_{ce}(e(\bm{s}_{i}), \bm{y}_{i})~~.~~
\end{equation}

\begin{table*}[!htp]
\caption{The comparison between our method and other SOTA methods on ImageNet and CUB-200 datasets.}
\centering	
\setlength{\tabcolsep}{3pt}
\begin{tabular}{c|ccc|ccccc}
\hline
~  & \multicolumn{3}{c|}{ImageNet dataset} & \multicolumn{5}{c}{CUB-200 dataset}\\
Method & Top-1 Loc & GT-known & BoxAccV2 & Top-1 Loc & GT-known & BoxAccV2 & pIoU & PxAP\\
\hline
CAM~\cite{CAM}  
& 51.81 & 64.72 & 62.69
& 65.80 & 72.47 & 68.28 & 47.60 & 66.78
\\
HAS~\cite{HAS}  
& 51.61 & 64.42 & 62.40
& 51.28 & 70.99 & 64.50 & 49.82 & 71.32
 \\
ACoL~\cite{ACOL} 
& 45.07 & 64.21 & 61.87 
& 42.53 & 70.09 & 61.22 & 41.56 & 56.78
 \\
SPG~\cite{SPG}  
& 46.62 & 63.71 & 61.36 
& 53.02 & 68.82 & 60.36 & 44.97 & 61.20
\\
ADL~\cite{ADL}  
& 49.81 & 64.67 & 62.75
& 44.30 & 63.31 & 56.61 & 42.29 & 56.59
\\
CutMix~\cite{CUTMIX}  
& 50.64 & 63.27 & 61.51
& 69.37 & 81.10 & 68.64 & 45.89 & 64.64
 \\
CAAM~\cite{CAAM}  
& 52.36 & 67.89 & - 
& 64.70 & 77.35 & - & - & -
\\
DGL~\cite{DGL}  
& 43.41 & 67.52 & - 
& 60.82 & 76.65 & - & - & -
\\
I2C~\cite{I2C}  
& 54.83 & 68.50 & -
& - & - & - & - & -
 \\
ICLCA~\cite{ICLCA}  
& 48.40 & 67.62 & 65.15
& 56.10 & 72.79 & 63.20 & - & -
 \\
\underline{PSOL}~\cite{PSOL}  
& 53.98 & 65.54 & -
& 70.68 & - & - & - & -
 \\
\underline{SEM}~\cite{SEM}  
& 53.84 & 67.00 & -
& - & - & - & - & -
 \\
\underline{FAM}~\cite{FAM}  
& 54.46 & 64.56 & -
& \textbf{73.74} & \textbf{85.73} & - & - & -
 \\
PAS~\cite{PAS}  
& 49.42 & 62.20 & 64.72 
& 59.53 & 77.58 & 66.38 & - & -
\\
IVR~\cite{IVR}  
& - & - & 65.57
& - & - & \textbf{71.23} & - & -
\\
\hline
Ours  
& 43.26 & \textbf{70.27} & \textbf{68.23}
& 62.40 & 81.83 & 69.87 & \textbf{56.18} & \textbf{74.70}
 \\
Ours$^*$  
& \textbf{55.84} & \textbf{70.27} & \textbf{68.23}
& 66.65 & 81.83 & 69.87 & \textbf{56.18} & \textbf{74.70}
 \\
\hline
\end{tabular}
\begin{flushleft}
$*$ \textit{Scores in \textbf{bold style} indicate the best. Methods in \underline{underline style} mean the generated localization maps are class-agnostic.}
\end{flushleft}
\label{tab:ImageNet_CUB}
\end{table*}

Based on these two domains and the image-level mask $\bm{y}$, the proposed target sample assigner selects representative samples of the three target subsets, $\mathbb{T}^{u}$ (noted by gray round), $\mathbb{T}^{f}$ (noted by peach round), and $\mathbb{T}^{t}$ (noted by pink round). Then, the samples of $\mathbb{T}^{u}$ are utilized to calculate the Universum regularization $\mathcal{L}_{u}$ with Eq.~\ref{eq:Universum}. The samples of the other two subsets and source domain (noted by pink square) are fed into existing UDA methods to align features distribution between source and target domain. Here, we adopt the MMD~\cite{MMD} as the UDA method:
\begin{equation}
\begin{split}
\mathcal{L}_{d}&(\mathbb{S} \cup \mathbb{T}^{f}, \mathbb{T}^{t}) = - \frac{\sum_{\bm{s}_{i} \in \mathbb{S} \cup \mathbb{T}^{f}, \bm{t}_{j} \in \mathbb{T}^{t}} 2*h(\bm{s}_{i}, \bm{t}_{j})}{|\mathbb{S} \cup \mathbb{T}^{f}|*|\mathbb{T}^{t}|} \\
&+ \frac{ \sum_{\bm{s}_{i}, \bm{s}_{j} \in \mathbb{S} \cup \mathbb{T}^{f}} h(\bm{s}_{i}, \bm{s}_{j})}{|\mathbb{S} \cup \mathbb{T}^{f}|^{2}}
+ \frac{\sum_{\bm{t}_{i}, \bm{t}_{j} \in \mathbb{T}^{t}} h(\bm{t}_{i}, \bm{t}_{j})}{|\mathbb{T}^{t}|^{2}}
\end{split}
\end{equation}
\noindent where $h(\cdot)$ is the gaussian kernel. Note that it is convenient to change the UDA or WSOL method in our DA-WSOL.

In the test processing, localization maps can be generated by directly feeding the target feature into the source-trained estimator, \textit{i.e.} $\bm{Y}^{*} = e ( f(\bm{X}) )$, whose column vector also represents the classification score of a target sample.

\section{Experiment}
In this section, we first introduce the settings and training details of our experiments. Then the results on three datasets are given to evaluate the effectiveness of our method. Next, ablation studies are conducted to explore the effect of different settings of our method. Finally, our defect and fail cases are also discussed to inspire future works. 
\subsection{Settings}
Our DA-WSOL adopted the CAM with ResNet50~\cite{RESNET} as the basic WSOL method and MMD~\cite{MMD} as the UDA approaches unless otherwise stated. The batch size $32$ was adopted, and the hyper-parameter $n$ was set $32$. Random cropped and random flip with size $224 \times 224$ were adopted as the augmentation. The SGD optimizer with weight decay $1e$-$4$ and momentum $0.9$ was utilized for training. 

Experiments were conducted on three widely-used WSOL benchmarks based on Pytorch toolbox~\cite{PYTORCH} with Intel Core i9 CPU and an Nvidia RTX 3090 GPU:

\noindent \textbf{ImageNet dataset~\cite{IMAGENET}} contains 1.3 million images of 1,000 classes, where 50,000 images with bounding box annotation served as the test set, and others were used as the training set. For the ImageNet dataset, the initial learning rate $1e$-$5$ was set to train our DA-WSOL for total $10$ epochs, which was divided by $10$ every $3$ epochs. Hyper-parameter $\lambda_{1}$ and $\lambda_{2}$ were set as $0.3$ and $3$, respectively.

\noindent  \textbf{CUB-200 dataset~\cite{CUB}} contains 11,788 images with 200 fine-grain classes of birds, where 5,794 images that have both pixel-level masks and bounding-box annotations were used as the testing set. Additionally, 1,000 extra images annotated by Junsuk~\cite{EVAL} were used as the validation set. The initial learning rate $1e$-$3$ was set for the SGD to train our DA-WSOL 50 epochs on this dataset. The learning rate was divided by 10 after training 30 epochs. The two hyper-parameters of DAL loss were set as $\lambda_{1}=0.3$ and $\lambda_{2}=2$.

\noindent \textbf{OpenImages dataset~\cite{EVAL}} contains 37,319 images of 100 classes, where the pixel-level mask annotations were released for 2,500 validation images and 5,000 test images. The rest 19,819 images served as the training set to train our DA-WSOL total $10$ epochs with the initial learning rate $1e$-$3$. The learning rate was divided by $10$ every $3$ epochs. Hyper-parameter $\lambda_{1}$ and $\lambda_{2}$ were set as $0.2$ and $3$.

The localization accuracy is evaluated by Top-1 localization accuracy (Top-1 Loc), ground truth known localization accuracy (GT-known), and maximal box accuracy (BoxAccV2)~\cite{EVAL} based on bounding boxes annotations. While, if the pixel-level annotations were available, peak intersection over union (pIoU) and the pixel average precision (PxAP)~\cite{EVAL} were adopted as the evaluation metric.

Except for our method (noted by Ours), we implemented six WSOL methods to fair compare the performance, including CAM~\cite{CAM}, HAS~\cite{HAS}, ACoL~\cite{ACOL}, SPG~\cite{SPG}, ADL~\cite{ADL}, and CutMix~\cite{CUTMIX}. Results of the other SOTA methods were cited from their papers. In addition, we also implemented a two-stage version (noted by Ours$^{*}$) that uses an additional stage to output the classification score.

\subsection{Results}
Table.~\ref{tab:ImageNet_CUB} shows the corresponding results of different WSOL methods and our proposed method on ImageNet and CUB-200 datasets with ResNet50 backbone. It shows that our method outperforms all those methods on all metrics of the large-scale ImageNet dataset. Specifically, benefited from eliminating the domain discrepancy between training and test process, our method achieves 1.77\% (about 885 images) and 2.66\% higher scores than the best of others respectively on GT-known and the BoxAccV2 metric. Though the classification accuracy is weakened due to the side-effect of the discrepancy elimination, adding a classification stage for generating classification results (Ours$^{*}$) can solve this defect and makes our method still outperform others on the Top-1 localization scores. As for the fine-grained CUB-200 dataset, methods that only generate class-agnostic localization results commonly have better results than others. This is because the class-agnostic localization can only focus on catching birds rather than different types of birds, which contributes to their higher performance. Though lower than this type of methods, our method achieves the highest GT-known, pIoU and PxAP scores among the methods that generate class-aware localization maps as ours.

In addition, results on the recently proposed OpenImages dataset are shown in Table.~\ref{tab:open}. Though the OpenImages dataset is more challenging because of its richer content and finer pixel-level evaluation, our method outperforms the existing methods by a large margin. Specifically, our method obtains 49.68\% pIoU and 65.42\% PxAP, which are 7.48\% and 4.52\% higher than the best of others. Two aspects profit this noticeable improvement. Firstly, the richer context of the OpenImages dataset provides more various samples, which helps estimate the feature distribution between source and target domain. This enhances the activation of the localization map on object locations. Secondly, the Universum regularization of our method also aligns the decision boundary to Universum samples, \textit{i.e.} background locations. This restrains the activation of the background location and promotes the pixel-level evaluation metrics. Moreover, our highest pixel-level evaluation metrics (pIoU and PxAP) of the CUB-200 dataset shown in Table.~\ref{tab:ImageNet_CUB} also prove this trait. 
\begin{table}[!htp]
\caption{Comparing with SOTA method on OpenImage dataset}
\centering	
\begin{tabular}{c|cc|cc}
\hline
~ & \multicolumn{2}{c|}{ResNet50} & \multicolumn{2}{c}{InceptionV3}\\
~ & pIoU & PxAP& pIoU & PxAP \\
\hline
CAM~\cite{CAM}  & 42.95 & 58.19 & 47.30 & 62.66 \\
HAS~\cite{HAS}   & 41.92 & 55.10 & 42.31 & 58.53 \\
ACoL~\cite{ACOL}  & 41.68 & 56.37 & 41.11 & 55.69 \\
SPG~\cite{SPG}  & 41.79 & 55.76 & 45.58 & 61.77 \\
ADL~\cite{ADL}  & 42.05 & 55.02 & 45.67 & 61.52 \\
CutMix~\cite{CUTMIX}  & 42.73 & 57.57 & 46.18 & 61.18 \\
PAS~\cite{PAS}  & - & 60.90 & - & 63.30 \\
IVR~\cite{IVR}  & - & 58.90 & - & 64.08 \\
\hline
Ours & \textbf{49.68} & \textbf{65.42} & \textbf{48.01} & \textbf{64.46} \\
\hline
\end{tabular}
\label{tab:open}
\end{table}
\begin{figure}[!htp]
\centering
\includegraphics[width=0.49\textwidth]{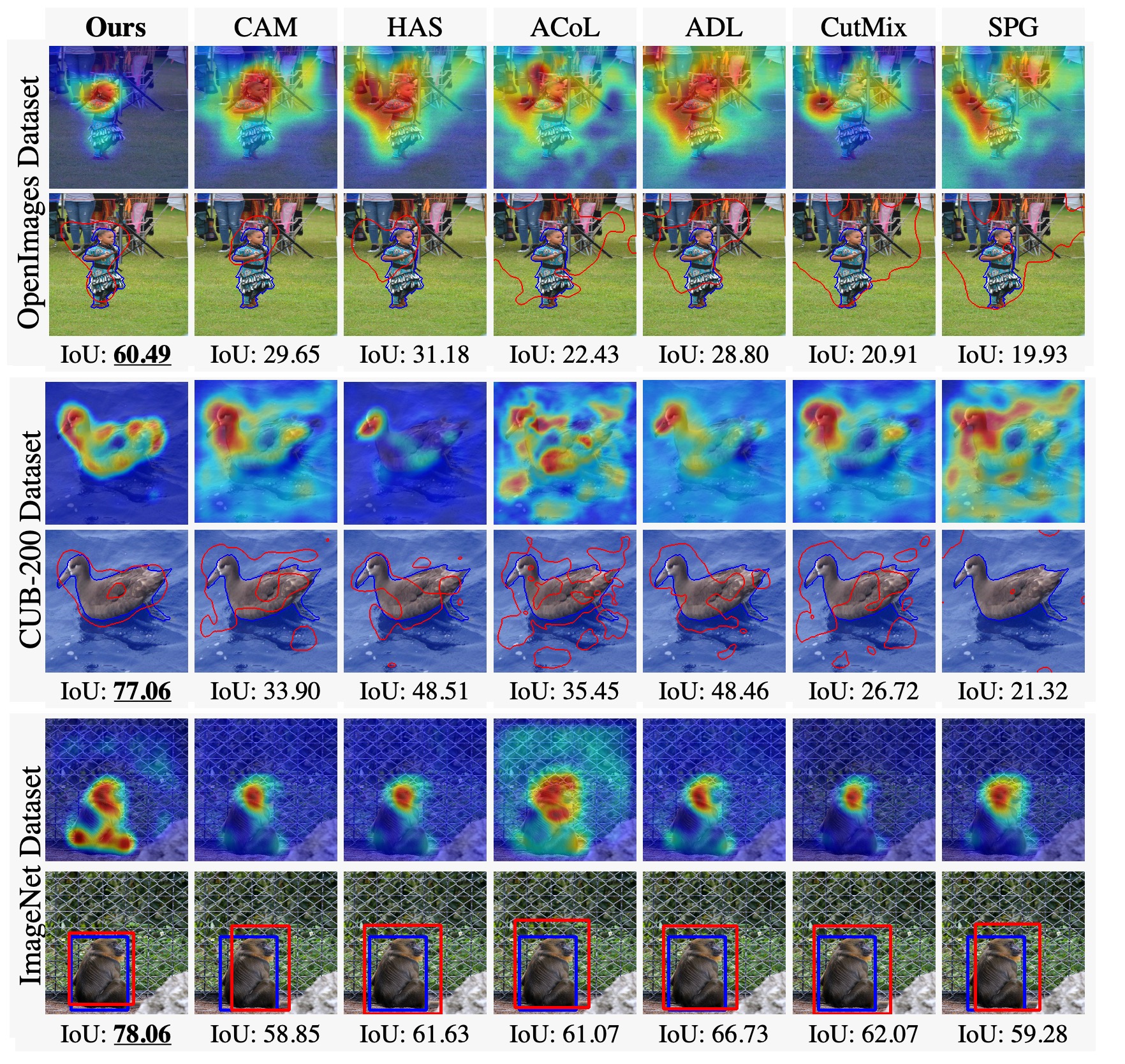}
\caption{The localization maps and boxes (or masks) for different WSOL methods under their optimal background threshold.}
\label{fig:vis}
\end{figure}

Except for the quantitive results,  Fig.~\ref{fig:vis} also visualizes some localization results. It can be seen that our method activates more object locations than other methods. This is because our method can reduce the feature distribution between the image domain and pixel domain, which pushes the feature of indiscriminative pixels (the body of duck and orangutan) to the image-level feature and makes them better activated by the estimator. This also helps to purify the uncorrelated objects (the adult or the chair) for the localization map of a certain class (the child). Moreover, benefiting from our Universum regularization, the localization map of our method also has the lowest interference on background locations (the water or the ground). These visualization aspects also qualitatively reflect the effectiveness of our pipeline that adopts domain adaption to assist WSOL. 

\subsection{Ablation Studies}
Ablation studies were conducted on the OpenImages dataset for our proposed DA-WSOL pipeline. Firstly, we explored the impact of our sampling strategy and DAL loss, \textit{i.e.} the TSA module, the domain adaption loss $\mathcal{L}_{d}$ and the Universum regularization $\mathcal{L}_{u}$. Corresponding results are shown in Table.~\ref{tab:ab_loss}. Compared with the baseline that only adopts classification loss, simply adding existing DA loss to align features will cause performance drops, because the features of Universum samples are also aligned with the features of objects, making the classifier discern background as objects. Under such condition, features of all target samples are messily distributed, which also disables the proposed TSA to effectively discern the Universum samples and assign them into $\mathbb{T}^{u}$. Thus, even though adopting TSA to balance the sample number between source and target domain for $\mathcal{L}_{d}$, the adaption loss still not fully take effect. However, when utilizing $\mathcal{L}_{u}$ to push feature of Universum samples (background location) into the decision boundary, TSA can better discern them from other target samples (object locations), which obviously promotes 2.22\% in pIoU and 3.31\% in PxAP. Thus, adopting the whole DAL loss can improve the performance to a great content compared with the baseline (6.73\% higher pIoU and 7.24\% higher PxAP).

\begin{table}[!htp]
\caption{Ablating parts of our method with DAs on OpenImages.}
\centering	
\begin{tabular}{cccc|cc|cc}
\hline
\multicolumn{4}{c|}{Settings}& \multicolumn{2}{c|}{MMD} & \multicolumn{2}{c}{DANN}\\
$\mathcal{L}_{c}$ & $\mathcal{L}_{d}$ & $\mathcal{L}_{u}$ & TSA & pIoU & PxAP & pIoU & PxAP\\
\hline
$\checkmark$ & ~ & ~ & ~ & 42.95 & 58.19 & 42.95 & 58.19 \\
$\checkmark$ & $\checkmark$ & ~ & ~ & 42.88 & 57.39 & 43.29 & 58.19  \\
$\checkmark$ & $\checkmark$ & ~ & $\checkmark$ & 43.24 & 58.10 & 43.67 & 58.77\\
$\checkmark$ & ~ & $\checkmark$ & $\checkmark$ & 45.17 & 61.50 &45.17 & 61.50  \\
$\checkmark$ & $\checkmark$ & $\checkmark$ & $\checkmark$ & 49.68 & 65.42 & 46.71 & 63.26\\
\hline
\end{tabular}
\label{tab:ab_loss}
\end{table}

To show the generalization of our DA-WSOL pipeline, results with the InceptionV3 on the OpenImages dataset are also given in Table.~\ref{tab:open}. It can be seen that our method also effectively enhances the performance of CAM. Moreover, we also adopted different UDA methods and WSOL methods for our DA-WSOL pipeline. Specifically, except for the MMD, we also utilized the adversarial learning based UDA method DANN~\cite{DANN} (structure shown in Appendix) to enhance three different WSOL methods, including the CAM~\cite{CAM}, HAS~\cite{HAS}, CutMix~\cite{CUTMIX} and ADL~\cite{ADL}. Results in Table.~\ref{tab:ab_uda} reflect that the performance of all those WSOL can be enhanced by UDA with our DA-WSOL pipeline.
\begin{table}[!htp]
\caption{Adopting different UDA and WSOL methods. }
\centering	
\setlength{\tabcolsep}{3pt}
\begin{tabular}{c|c|cc|cc}
\hline
~ & UDA & pIoU & PxAP & Train & Test\\
\hline
~ & - & 42.95 & 58.19 & 13.84 & 69.77 \\
CAM & MMD & 49.68$^{\textcolor{red}{\uparrow}{6.73}}$ & 65.42$^{\textcolor{red}{\uparrow}{7.23}}$ & 25.25 & 70.84 \\
~ & DANN & 46.71$^{\textcolor{red}{\uparrow}{3.76}}$ & 63.26$^{\textcolor{red}{\uparrow}{5.07}}$ & 19.36 & 70.48 \\
\hline
~ & - & 41.92 & 55.10 & 13.75 & 70.13 \\
HAS & MMD & 49.25$^{\textcolor{red}{\uparrow}{7.33}}$ & 64.15$^{\textcolor{red}{\uparrow}{9.05}}$ & 20.52 & 70.69 \\
~ & DANN & 46.41$^{\textcolor{red}{\uparrow}{4.49}}$ & 62.48$^{\textcolor{red}{\uparrow}{7.38}}$ & 19.41 & 71.17 \\
\hline
~ & - & 42.73 & 60.64 & 11.17 & 69.24 \\
CutMix & MMD & 49.32$^{\textcolor{red}{\uparrow}{6.59}}$ & 64.68$^{\textcolor{red}{\uparrow}{4.04}}$ & 31.23 & 71.05 \\
~ & DANN & 48.55$^{\textcolor{red}{\uparrow}{5.82}}$ & 64.08$^{\textcolor{red}{\uparrow}{3.44}}$ & 28.61 & 70.40 \\
\hline
~ & - & 42.05 & 55.02 & 8.87 & 69.32 \\
ADL & MMD & 43.40$^{\textcolor{red}{\uparrow}{1.35}}$ & 60.17$^{\textcolor{red}{\uparrow}{5.15}}$ & 20.38 & 71.38 \\
~ & DANN & 43.22$^{\textcolor{red}{\uparrow}{1.17}}$ & 60.64$^{\textcolor{red}{\uparrow}{5.62}}$ & 16.90 & 70.26 \\
\hline
\end{tabular}
\label{tab:ab_uda}
\begin{flushleft}
$*$ \textit{Train/Test metric is the training/test time (ms per image).}
\end{flushleft}
\end{table}

\subsection{Limitations}
Though the proposed DA-WSOL pipeline helps to engage DA methods to enhance WSOL and refreshes the SOTA localization performance on the ImageNet and OpenImages datasets, some limitations of our method should also be concerned. Firstly, adopting DA for WSOL negatively affects the strength of the estimator on the source domain, which is also required for WSOL to choose the localization map of different classes. This defect weakens the image classification accuracy when the classification task is challenged and causes our low Top-1 metric on the ImageNet and CUB-200 datasets. Moreover, our TSA adopts the time-consuming K-Means cluster to sample the target samples and update the anchors of different subsets. Though it does not influence the model size and the time of the test process, the training time is increased compared with the WSOL method adopted by our DA-WSOL pipeline. We hope future works can solve these problems to enhance our work.

\section{Conclusion}
This paper provides a novel perspective that models WSOL as a DA task and proposes a DA-WSOL pipeline to assist WSOL with DAs. Our method uses a target sampling strategy to assign target samples into three subsets, which are then adopted by DAL loss to concern specificities. Experiments indicate our method outperforms SOTA methods on multi datasets and can generalize to various baselines.

\section{Acknowledgements} This work was supported by the Beijing Natural Science Foundation under Grant Z210008, the Shenzhen Science and Technology Program under Grant 1210318663, and the Development of Key Technologies and Equipment in Major Science and Technology Infrastructure in Shenzhen.

\clearpage
{\small
\bibliographystyle{ieee_fullname}
\bibliography{egbib}
}

\end{document}